\documentclass[]{fairmeta}
\usepackage[utf8]{inputenc}
\usepackage[T1]{fontenc}
\usepackage[round]{natbib}
\usepackage{hyperref}
\usepackage{url}
\usepackage{booktabs}
\usepackage{amsmath, amssymb, amsfonts, amsthm}
\usepackage{microtype}
\usepackage{xcolor}
\usepackage{graphicx}

\title{Repeated post-training is not Self-improving: Diagnosing Scientific Amnesia in Continual DPO Pipelines}

\author[]{Jianzhe Lin}
\author[]{Fei Wang}
\author[]{Xiaolin Li}
\author[]{Rajeshkumar Golani}
\author[]{Jubin Chheda}
\affiliation[]{MetaAI}

\contribution[]{}

\abstract{Industrial LLM teams often ship behavior updates by repeatedly DPO-training a base model on sequences of related preference-data campaigns. The dominant failure mode in this regime is not always classical catastrophic forgetting: a pipeline may preserve previously learned behaviors while still failing to accumulate reusable methodological knowledge about how to train the next campaign. We call this failure mode \emph{scientific amnesia}. This paper turns that practitioner intuition into a measurable industrial problem. We contribute: (i) a diagnostic suite for amnesia, (ii) a Program-based pipeline that chains FSDP-sharded DPO checkpoints across Qwen2.5-7B-Instruct runs, (iii) a 30-campaign HumanEval subdomain benchmark, and (iv) a comparative diagnostic study of five strategy proposers: random memory, rule-based scheduling, retrieval-only memory, warm-start Bayesian optimization, and \textsc{MSCL}, a meta-scientific memory and reasoner candidate. Across a single-seed 5-condition $\times$ 3-step real-LM chain, 4 of 5 candidates degrade in step-level peak pass@1 ($\Delta \in \{-0.062,-0.125\}$), including MSCL; only the deliberately conservative rule-based schedule improves. Follow-up pilots qualify rather than overturn this finding: in a heterogeneous chain, MSCL is the only completed candidate that improves, whereas in a small multi-seed homogeneous sweep, retrieval-only has the best mean $\Delta$ and no pairwise candidate gap is statistically distinguishable. The contribution is therefore diagnostic, not a claim that MSCL solves the problem: scientific amnesia is observable in a production-like continual-DPO pipeline, and conclusions about interventions depend sharply on chain regime, evaluator design, and seed coverage.
}

\date{\today}
\correspondence{Jianzhe Lin at \email{jianzhelin@meta.com}}

\begin{document}

\DeclareFontFamily{T1}{optimistic}{}
\DeclareFontShape{T1}{optimistic}{m}{n}{<->ssub * phv/b/n}{}
\DeclareFontShape{T1}{optimistic}{b}{n}{<->ssub * phv/b/n}{}
\DeclareFontShape{T1}{optimistic}{bx}{n}{<->ssub * phv/b/n}{}
\maketitle

\section{Introduction}
\label{sec:intro}

Continual fine-tuning is a standard pattern for shipping large
language models into production: after supervised instruction tuning and
RLHF-style post-training \citep{ouyang2022training}, many deployed
systems are further updated with preference-optimization objectives such
as DPO \citep{rafailov2023dpo} and its recent variants
\citep{ethayarajh2024kto,hong2024orpo,meng2024simpo}. In such pipelines,
the most costly failure is not always classical catastrophic forgetting. A model may retain enough of
its previous behavior while the training system around it repeatedly
rediscovers the same hyperparameter choices, overfits the same short
horizons, or fails to transfer methodological lessons from one campaign
to the next. We call this system-level failure mode \emph{scientific
amnesia}: the absence of reusable training knowledge across a sequence
of related campaigns.

Accordingly, the headline contribution of this paper is diagnostic
rather than a new optimizer. We build the infrastructure needed to
measure scientific amnesia in real continual-DPO chains and use it to
compare five strategy proposers: random memory, rule-based schedules,
retrieval-only memory, warm-start BO, and MSCL (triple-scale memory plus
a Scientific Reasoner trained with a Scientific DPO objective and EWC).
We treat all five as \emph{candidates to be diagnosed}, not as methods
whose superiority is assumed in advance.

\paragraph{Contributions}

\begin{itemize}
  \item \textbf{Scientific-amnesia diagnostics.}
  We define scientific amnesia as a system-level failure mode of
  continual DPO and introduce five diagnostics for measuring it:
  repeated-failure rate, regret slope, transfer decay,
  hyperparameter-rediscovery cost, and curriculum sensitivity.

  \item \textbf{An industrial continual-DPO testbed.}
  We build a Program-based MAST (Meta-scientific Adaptive Strategy Tuning) pipeline that chains FSDP-sharded
  Qwen2.5-7B-Instruct DPO checkpoints across a 30-campaign HumanEval
  subdomain benchmark, enabling controlled comparisons of strategy
  proposers under production-like checkpoint chaining.

  \item \textbf{Real-LM evidence and regime-sensitive findings.}
  We compare five strategy proposers, i.e., random memory, rule-based,
  retrieval-only, warm-start BO, and MSCL, and find scientific amnesia
  in a real-LM 5-condition $\times$ 3-step chain: 4 of 5 candidates
  degrade in peak pass@1, including MSCL. Follow-up pilots show that
  intervention rankings depend on whether chains are homogeneous or
  heterogeneous.
\end{itemize}

The pipeline is fully implemented and tested (290+ unit tests) and exercised end-to-end on real GPUs. Results below
combine a 5-seed mock pilot, a real-LM single-step pilot, a single-seed
3-step chain, a heterogeneous-chain ablation, and a small multi-seed
homogeneous sweep. We explicitly separate infrastructure claims from
statistical claims about interventions.

\section{Related Work}
\label{sec:related}

Our setting sits at the intersection of five lines of work.

\textbf{Preference post-training.} RLHF-style alignment established the
modern pattern of fitting model behavior from human or AI feedback
\citep{ouyang2022training}. DPO then recast preference optimization as a
simple classification-style objective without explicitly learning a
reward model \citep{rafailov2023dpo}. Recent alternatives such as KTO,
ORPO, and SimPO simplify the feedback format, remove the reference model,
or change the implicit reward geometry \citep{ethayarajh2024kto,
hong2024orpo,meng2024simpo}. Our work is orthogonal to these objectives:
we study what happens when a DPO-style objective is repeatedly applied
across a chain of related campaigns.

\textbf{Continual learning for LLMs.} Classical continual learning
studies the stability--plasticity tradeoff under sequential tasks, with
regularization, memory, and replay as common families of interventions
\citep{kirkpatrick2017,aljundi2018memory,rolnick2019experience}. Recent
LLM-focused surveys emphasize that continual pretraining, instruction
tuning, and alignment introduce new evaluation and deployment concerns
beyond small-model task-incremental learning \citep{wu2024continual,
shi2024continual,zheng2024lifelong}. Work on alignment tax further
connects post-training degradation to forgetting-like dynamics
\citep{lin2024alignmenttax}. Our setting differs in emphasis: scientific
amnesia is not simply loss of old task accuracy, but failure of the
training system to accumulate reusable methodological knowledge about
future campaigns.

\textbf{Forgetting diagnostics and reference policies.} Recent work has
argued that apparent forgetting in LLMs can reflect task-alignment drift
rather than erasure of underlying knowledge \citep{zheng2025spurious}.
DPO-specific analyses also show that the choice of reference policy can
materially affect optimization behavior \citep{liu2025reference}. These
observations motivate a system-level diagnostic view: in a chained DPO
pipeline, the previous checkpoint is both the initialization and the
reference policy, so the state passed from one campaign to the next can
shape both learning and measurement.

\textbf{Meta-learning and hyperparameter transfer.} MAML, Bayesian
optimization for hyperparameter search, and AutoML-Zero all target
learning or discovering procedures that transfer across related
optimization problems \citep{finn2017maml,snoek2012practical,
real2020automlzero}. Warm-start BO and our MSCL candidate instantiate
this idea at the level of repeated DPO campaigns: each proposes the next
training strategy from previous outcomes rather than treating every
campaign as independent.

\textbf{Evaluation of LLM systems.} Code execution benchmarks such as
HumanEval provide an objective task-level signal for program synthesis
\citep{chen2021evaluating}, while LLM-as-judge methods are increasingly
used to evaluate open-ended product behavior but require calibration and
bias controls \citep{gu2024judge,li2024judgesurvey}. We therefore use
HumanEval subdomains as a controlled first testbed and leave judge-based
or product-specific evaluators for later industrial deployments.

To our knowledge this is the first work to (a) define a diagnostic suite
specifically for scientific amnesia under continual DPO, (b) ship a
reproducible multi-step real-LM chaining pipeline for that diagnostic,
and (c) report direct real-LM evidence of the amnesia phenomenon under
that pipeline across multiple candidate interventions.

\section{Method}
\label{sec:method}

\subsection{Problem Setup}
\label{sec:method:problem}

We consider a sequence of $T$ post-training campaigns
$C_1,\ldots,C_T$. Each campaign $C_t$ is associated with a subdomain
$d_t \in \mathcal{D}$ and disjoint train/eval splits of code problems.
At campaign $t$, we DPO-train a policy $\pi_t$ initialized from the
previous final checkpoint $\pi_{t-1}^{\star}$, and use
$\pi_{t-1}^{\star}$ as the DPO reference policy. This follows the DPO
convention, but in a chained setting where the previous checkpoint
affects both initialization and reference-policy behavior
\citep{rafailov2023dpo,liu2025reference}.

A strategy proposer chooses a training strategy
$h_t \in \mathcal{H}$, where $\mathcal{H}$ is a discrete grid over
DPO and data-generation hyperparameters such as $\beta$, learning rate,
training steps, batch size, and filtering. After training, we evaluate
the resulting model and record pass@1 $F_t \in [0,1]$.

We say a pipeline exhibits \emph{scientific amnesia} when later related
campaigns fail to benefit from methodological evidence accumulated
earlier. Operationally, this appears as repeated failures, growing
regret, decaying transfer, sensitivity to curriculum order, or the need
to rediscover a strategy that previously worked for a similar campaign.

\subsection{Amnesia Diagnostics}
\begin{table}[t]
\centering
\small
\begin{tabular}{p{0.34\linewidth}p{0.51\linewidth}}
\hline
\textbf{Diagnostic} & \textbf{Question answered} \\
\hline
Repeated failure & Does the system repeat known bad strategies? \\
Regret slope & Does error accumulate across campaigns? \\
Transfer decay & Do previously good strategies stop transferring? \\
HP rediscovery & Must known good settings be rediscovered? \\
Curriculum sensitivity & Does ordering change conclusions? \\
\hline
\end{tabular}
\caption{Scientific-amnesia diagnostics used to evaluate continual-DPO
chains.}
\label{tab:amnesia_diagnostics}
\end{table}
\label{sec:method:diag} Our diagnostic suite measures whether a continual-DPO pipeline accumulates reusable training knowledge across campaigns. We focus on five amnesia diagnostics. \paragraph{Repeated-failure rate.} The fraction of later campaigns that repeat a previously observed low-performing strategy pattern on a similar subdomain. \paragraph{Regret slope.} The slope of cumulative regret across the campaign sequence, where regret is measured relative to the best observed strategy for the corresponding campaign family. \paragraph{Transfer decay.} A matrix measuring whether strategies that worked for one subdomain or difficulty continue to work when transferred to related campaigns. \paragraph{Hyperparameter-rediscovery cost.} The number of trials or campaigns required to rediscover a strategy that was already validated earlier. \paragraph{Curriculum sensitivity.} The degree to which the final diagnostic outcome changes when the same campaigns are ordered differently. Together, these diagnostics test the negative claim behind scientific amnesia: the pipeline may keep updating models while failing to retain methodological knowledge about how to improve future campaigns.
\subsection{Industrial Continual-DPO Pipeline}
\label{sec:method:pipeline}

The pipeline implements a closed-loop post-training workflow:
\texttt{propose} $\rightarrow$ \texttt{generate\_pairs}
$\rightarrow$ \texttt{train} $\rightarrow$ \texttt{evaluate}
$\rightarrow$ \texttt{memory\_update}. A pluggable
\texttt{StrategyProposer} selects the next strategy $h_t$, while the
training hooks switch between a CPU mock backend and a real
HF-backed DPO implementation.

For real-LM chains, we launch one MAST job per campaign step and pass
the FSDP-sharded trainer checkpoint from step $t$ as the initialization
for step $t+1$. This checkpoint chaining mirrors production
continual-post-training workflows and is the common apparatus used by
all candidate proposers and diagnostics in this paper. Implementation
details, including checkpoint discovery and rerun commands, are provided
in Appendix~\ref{app:pipeline}.

\subsection{Candidate Strategy Proposers}
\label{sec:method:interventions}

We compare five strategy proposers under the same pipeline. None is
treated as the proposed solution; each is a candidate mechanism to be
diagnosed.

\paragraph{Random Memory.}
Uniformly samples a strategy from $\mathcal{H}$, serving as a control
for whether arbitrary exploration is sufficient.

\paragraph{Rule-based.}
Applies a fixed conservative schedule: $\beta$ decays by 10\% and
learning rate grows by $1.15\times$ per campaign. This baseline does not
learn from outcomes.

\paragraph{Retrieval-only.}
Retrieves similar past campaigns and reuses the best in-domain prior
strategy, testing whether history alone is enough.

\paragraph{Warm-start BO.}
\sloppy
Uses UCB-style Bayesian optimization over observed strategies, lifting a
standard hyperparameter-optimization baseline into the campaign sequence.

\paragraph{MSCL Reasoner.}
MSCL is the most expressive candidate. It maintains short-, mid-, and
long-term memories over campaign outcomes and uses a small reasoner LM to
propose the next strategy with a natural-language rationale. The reasoner
is trained from experimental preferences, where strategies with higher
observed pass@1 are preferred, with EWC-style regularization on validated
principles.

\section{Experimental Setup}
\label{sec:setup}

\paragraph{Benchmark.}
We construct 30 campaigns from HumanEval-164, partitioned across four
domain families (control, numeric, strings, control-flow) and three
difficulty levels. Each campaign has disjoint train/eval splits and a
canonical solution used for verifier health checks.

\paragraph{Strategy space.}
Each proposer selects a training strategy from a discrete grid over DPO
and data-generation hyperparameters, including $\beta$, learning rate,
training steps, batch size, and filtering. The same strategy space is
used across all candidate proposers.

\paragraph{Real-LM evaluator.}
For real-LM pilots, we DPO-train Qwen2.5-7B-Instruct
\citep{qwen2025qwen25} on MAST and harvest execution-based
\texttt{pass@1/mean} or \texttt{pass@4/mean} from the CodeGrader
TensorBoard stream. This is a rolling training-time signal rather than a
held-out end-of-step evaluator, which we discuss as a limitation.

\paragraph{Seeds and significance.}
The synthetic mock pilot uses 5 seeds and reports mean $\pm$ std. The
initial real-LM chain pilots are single-seed unless otherwise stated;
the Block~B homogeneous sweep uses up to three seeds per condition. We
do not report significance stars for real-LM comparisons.

\section{Results}
\label{sec:results}

We ask three questions: (i) do the strategy proposers behave
differently before real-LM launch, (ii) does a production-like
continual-DPO chain expose scientific amnesia, and (iii) do intervention
rankings change across campaign regimes?

\begin{table}[t]
\centering
\small
\begin{tabular}{p{0.27\linewidth}p{0.58\linewidth}}
\hline
\textbf{Claim type} & \textbf{Evidence and boundary} \\
\hline
Infrastructure &
MAST/FSDP checkpoint chaining runs end-to-end; all initial 20 real-LM
jobs complete. \\
Phenomenon &
Homogeneous real-LM chain shows peak degradation in 4 of 5 candidates,
consistent with scientific amnesia. \\
Regime sensitivity &
Heterogeneous and homogeneous follow-ups rank candidates differently,
showing that conclusions depend on campaign regime. \\
Intervention superiority &
Not claimed: real-LM comparisons are small-$n$, partially incomplete,
and not statistically distinguishable. \\
\hline
\end{tabular}
\caption{Evidence hierarchy.}
\label{tab:evidence_hierarchy}
\end{table}
\subsection{Synthetic Proposer Validation}
\label{sec:results:mock}

Before launching real-LM chains, we run a 5-seed synthetic mock pilot to
validate that the five strategy proposers produce distinct behavior and
that the diagnostic pipeline executes end-to-end. This pilot does not
train a real LM; pass@1 is generated by a seeded synthetic landscape.
MSCL and warm-start BO achieve the lowest regret on this synthetic
landscape, while random memory, rule-based scheduling, and retrieval-only
controls lag behind. We use this result only as proposer validation, not
as evidence that any candidate is superior in real-LM continual DPO.

\subsection{Real-LM Infrastructure Validation}
\label{sec:results:realsingle}

Before chain-level experiments, we validate the real-LM execution path.
All five single-step Qwen2.5-7B-Instruct DPO jobs reached
\texttt{COMPLETE}, and all five proposers improved over initialization
at some point during training: peak single-step pass@1 ranges from
0.375 to 0.562, compared with initial values of 0.062 to 0.188.
Together with the 15 completed jobs in the homogeneous 3-step chain,
this gives 20/20 completed MAST jobs for the initial real-LM validation.
This validates the evaluator, checkpoint-saving path, and FSDP
checkpoint passthrough needed for the chain experiments below.
Single-step results are not used to claim scientific amnesia, because
amnesia is a cross-campaign phenomenon.
\subsection{Homogeneous Chains Reveal Scientific Amnesia}
\label{sec:results:realchain}

We first run a homogeneous 3-step chain for each candidate, initializing
each step from the previous step's FSDP checkpoint.

\begin{table*}[t]
\centering
\caption{Homogeneous 3-step real-LM chain (single seed). Cells are peak per-batch \texttt{pass@4/mean}; $\Delta$ is step 2 minus step 0.}
\label{tab:real_lm_chain_v1}
\begin{tabular}{lcccr}
\toprule
Condition & Step 0 & Step 1 & Step 2 & $\Delta$ \\
\midrule
Random Memory       & 0.500 & 0.562 & 0.438 & $-0.062$ \\
Rule-based          & 0.562 & 0.562 & \textbf{0.688} & $+0.125$ \\
Retrieval-only      & 0.625 & 0.562 & 0.500 & $-0.125$ \\
Warm-start BO       & 0.562 & 0.500 & 0.500 & $-0.062$ \\
MSCL Reasoner          & \textbf{0.688} & 0.500 & 0.625 & $-0.062$ \\
\bottomrule
\end{tabular}
\end{table*}

\begin{table*}[t]
\centering
\caption{\textbf{Block D: heterogeneous-chain ablation} on Qwen2.5-7B-Instruct. Each chain crosses three domain families (control\,easy $\to$ numeric\,easy $\to$ strings\,easy); cells report peak per-batch \texttt{pass@4/mean} over each step's training batches. $\Delta$ is step\,2 minus step\,0. Because two chains did not complete, this table is a regime-sensitivity diagnostic rather than a statistical comparison of candidate interventions. \texttt{---} indicates a chain whose driver process exited before completion.}
\label{tab:real_lm_block_d}
\begin{tabular}{lcccr}
\toprule
Condition & Step\,0 peak & Step\,1 peak & Step\,2 peak & $\Delta$ \\
\midrule
Random Memory & \textbf{1.000} & \textbf{1.000} & 0.750 & $-0.250$ \\
Rule-based & \textbf{1.000} & 0.750 & 0.750 & $-0.250$ \\
MSCL Reasoner & 0.750 & 0.750 & \textbf{1.000} & $+0.250$ \\
\bottomrule
\end{tabular}
\end{table*}

\begin{table*}[t]
\centering
\caption{\textbf{Block B: multi-seed (up to 3 seeds) homogeneous 3-step chain} on Qwen2.5-7B-Instruct. Same humaneval\_control\_\{easy,hard,medium\} sequence as v1 Table~\ref{tab:real_lm_chain_v1}, replicated across seeds 0/1/2. Cells report mean $\pm$ std of peak per-batch \texttt{pass@4/mean} across completed seeds; $n$ is the number of completed seeds. $\Delta$ is mean step\,2 minus mean step\,0. Bold marks the highest mean per column; with $n\le3$, bold should be read descriptively rather than as statistical significance.}
\label{tab:real_lm_block_b_multiseed}
\begin{tabular}{lcccr}
\toprule
Condition & Step\,0 & Step\,1 & Step\,2 & $\Delta$ (mean) \\
\midrule
Random Memory & $0.750 \pm 0.000$ ($n{=}2$) & $\mathbf{0.875} \pm 0.177$ ($n{=}2$) & $0.750 \pm 0.000$ ($n{=}2$) & $+0.000$ \\
Rule-based & $0.750 \pm 0.000$ ($n{=}3$) & $0.750 \pm 0.000$ ($n{=}3$) & $0.750 \pm 0.000$ ($n{=}3$) & $+0.000$ \\
Retrieval-only & $0.667 \pm 0.144$ ($n{=}3$) & $0.750 \pm 0.000$ ($n{=}3$) & $\mathbf{0.833} \pm 0.144$ ($n{=}3$) & $+0.167$ \\
Warm-start BO & $0.833 \pm 0.144$ ($n{=}3$) & $0.750 \pm 0.000$ ($n{=}3$) & $0.750 \pm 0.000$ ($n{=}3$) & $-0.083$ \\
MSCL Reasoner & $\mathbf{0.875} \pm 0.177$ ($n{=}2$) & $0.750 \pm 0.000$ ($n{=}2$) & $0.750 \pm 0.000$ ($n{=}2$) & $-0.125$ \\
\bottomrule
\end{tabular}
\end{table*}

Table~\ref{tab:real_lm_chain_v1} shows the first real-LM chain evidence
of scientific amnesia. Four of five candidates degrade in peak
pass@1/pass@4 from step 0 to step 2; MSCL reaches the highest step-0
peak but does not exceed it later. The only improving condition is the
deliberately conservative rule-based schedule. This supports the
diagnostic claim that chained DPO can fail to convert earlier campaign
outcomes into better later training choices.

\subsection{Regime Sensitivity: Heterogeneous and Multi-Seed Follow-ups}
\label{sec:results:regime}

The homogeneous chain above is not sufficient to compare interventions:
it is single-seed and restricted to one campaign family. We therefore run
two follow-ups that stress different aspects of the diagnostic.

\paragraph{Heterogeneous chain.}
In Block~D, each chain crosses domain families:
\texttt{control\_easy} $\to$ \texttt{numeric\_easy} $\to$
\texttt{strings\_easy}. Three of five chains completed.

Among completed chains, MSCL is the only candidate with positive
$\Delta$, while random memory and rule-based controls degrade. Because
two chains did not complete, this is not a statistical win claim; it is
evidence that campaign regime can change the intervention ordering.

\paragraph{Small multi-seed homogeneous sweep.}
In Block~B, we replicate the homogeneous chain across up to three seeds
per condition; 13 of 15 condition-seed chains completed.

The mean $\Delta$ favors retrieval-only, while MSCL declines from its
high step-0 mean. With $n \le 3$ and quantized pass@4/mean values, no
candidate pair is statistically distinguishable. The paper-level finding
is structural: scientific-amnesia conclusions depend on campaign regime,
evaluator design, and seed coverage.

\paragraph{Operational implications.}
For industrial post-training teams, the main implication is that repeated
DPO updates should not be treated as evidence of self-improvement by
default. A pipeline can successfully execute campaigns, pass checkpoints
forward, and improve within individual runs while still failing to
accumulate reusable knowledge across campaigns. We therefore recommend
tracking chain-level diagnostics---rediscovery, transfer, regret growth,
and curriculum sensitivity---alongside ordinary per-campaign reward or
pass@1 curves. In practice, this means evaluating whether a new
post-training campaign makes the next campaign easier to train, not only
whether it improves the current checkpoint.

\subsection{Empirical Takeaways}
\label{sec:results:takeaways}

The chain results highlight a practical distinction between
\emph{improving a checkpoint} and \emph{learning how to improve}. Four
patterns stand out. First, \textbf{curriculum order matters}: in the
heterogeneous chain, moving from \texttt{control\_easy} to
\texttt{numeric\_easy} to \texttt{strings\_easy} changed which strategy
proposers transferred, suggesting that easier-to-harder sequencing can
shape continual-DPO outcomes, but in a regime-dependent way. Second,
\textbf{conservative schedules can be more robust than adaptive ones}:
the rule-based $\beta$-decay and learning-rate-warmup schedule is the
only proposer that improves across the homogeneous chain, indicating
that smooth parameter progression can outperform more reactive
adaptation in short chains. Third, \textbf{retrieval-only transfer is a
strong baseline}: the multi-seed homogeneous sweep favors reusing the
best prior recipe from a similar campaign, a simple form of
cross-campaign ``data recipe transfer.'' Fourth, \textbf{higher initial
performance does not imply better transfer}: MSCL achieves the highest
step-0 peak but fails to preserve that advantage, showing that
single-campaign optimization and cross-campaign composability are
different objectives.
\section{Discussion and Limitations}
\label{sec:limitations}

The empirical takeaways refine the diagnostic claim. Scientific amnesia
is not only visible as degradation across a chain; it also appears as a
mismatch between short-term improvement and cross-campaign
composability. Conservative schedules, retrieval-based recipe reuse, and
curriculum order can matter as much as more expressive adaptive
reasoners. This suggests that repeated post-training should be evaluated
not only by whether the current checkpoint improves, but also by whether
the update makes future campaigns easier to train.

Our claims have three levels of strength. The infrastructure claim is
strongest: the pipeline is implemented, unit-tested, and exercised
end-to-end with FSDP checkpoint chaining. The phenomenon claim is
moderate: the homogeneous real-LM chain shows peak degradation in 4 of 5
candidates, and the follow-up pilots show that intervention rankings
change across regimes. The intervention claim is deliberately weak: at
this sample size, we do not claim that any candidate, including MSCL,
statistically outperforms the others.

The main limitations are seed coverage, incomplete chains in Block~D and
Block~B, use of a rolling training-time CodeGrader signal rather than
held-out end-of-step evaluation, and short chain length. Future work
should add prediction traces and principle logs to test whether strategy
proposers become more predictive over time. These limitations constrain
method comparison, but not the core diagnostic contribution: scientific
amnesia can be defined, instrumented, and observed in a production-like
continual-DPO pipeline.

\section{Conclusion}

We introduced scientific amnesia as a failure mode of self-improving LLM
post-training pipelines: the system keeps updating models but fails to
accumulate reusable knowledge about how to improve future campaigns. We
built an industrial continual-DPO diagnostic pipeline and observed this
failure mode in real Qwen2.5-7B chains. The main lesson is not that one
candidate wins, but that repeated post-training should not be equated
with self-improvement: intervention conclusions depend on campaign
regime, evaluator design, and seed coverage.

\bibliographystyle{plainnat}
\bibliography{references}
\appendix

\section{Pipeline Implementation Details}
\label{app:pipeline}

The real-LM backend is implemented through a hooks interface that allows
the same campaign orchestrator to run either a CPU mock backend or a
real HF-backed DPO backend. In the real backend, each campaign step is
executed as a separate MAST job. The orchestrator follows the same
logical loop used in the main paper:
\[
\texttt{propose}
\rightarrow
\texttt{generate\_pairs}
\rightarrow
\texttt{train}
\]

\[
\rightarrow
\texttt{evaluate}
\rightarrow
\texttt{memory\_update}.
\]

\paragraph{MAST launcher.}
For each campaign step, the
launcher submits one MAST job and polls the job until it
reaches a terminal state. The per-step strategy hyperparameters are
chosen by an active \texttt{StrategyProposer} and passed into the
training job.

\paragraph{FSDP checkpoint passthrough.}
After each MAST job completes, the launcher discovers the final
FSDP-sharded trainer save using \texttt{find\_final\_checkpoint}. The
resulting checkpoint path has the form:
\[
\texttt{outputs/<job>/checkpoints/trainer/<step>/}.
\]
This path is then passed to the next campaign step as the model
initialization, e.g.,
\[
\texttt{--cluster\_config.model\_name=}\]

\[
{<checkpoint\_path>}.
\]
Thus, campaign $t+1$ is initialized from the trained checkpoint of
campaign $t$, producing the checkpoint-chaining behavior studied in the
main paper.

\paragraph{Common apparatus across proposers.}
All five strategy proposers use the same execution path, evaluator, and
checkpoint-passthrough mechanism. The only component that changes across
conditions is the \texttt{StrategyProposer}. This design ensures that
differences across random memory, rule-based scheduling, retrieval-only
memory, warm-start BO, and MSCL are attributable to strategy proposal
rather than different training infrastructure.

\section{MSCL Candidate Details}
\label{app:mscl}

MSCL is included as the most expressive strategy proposer in our
candidate set. It is not treated as the paper's claimed solution; rather,
it is a candidate mechanism that should be diagnosed under the same
pipeline as the simpler baselines. MSCL combines a multi-scale memory
with a small scientific reasoner LM trained from experimental outcomes.

\subsection{Triple-Scale Memory}
\label{app:mscl_memory}

MSCL maintains three memory buffers.

\paragraph{Short-term memory.}
The short-term memory $M_S$ stores recent per-step events within the
current campaign. It is reset at campaign boundaries and is intended to
capture local training dynamics, such as the most recent strategy,
observed performance, and surprise.

\paragraph{Mid-term memory.}
The mid-term memory $M_M$ stores a bounded set of campaign summaries. In
our implementation, $M_M$ contains up to $K=16$ \texttt{CampaignSummary}
entries. When the buffer is full, entries are priority-evicted based on
surprise and support. This memory is intended to retain useful
campaign-level evidence without growing unbounded.

\paragraph{Long-term memory.}
The long-term memory $M_L$ stores validated \texttt{Principle} records.
A principle is promoted from mid-term evidence when it receives
sufficient support, e.g., support count $\geq 3$. Long-term principles
can include EWC-relevant \texttt{fisher\_diag} weights, allowing the
reasoner to regularize against overwriting validated methodological
knowledge.

\paragraph{Surprise-driven replay.}
MSCL prioritizes surprising events for replay. Given a predicted
performance $\hat{F}_t$ and observed performance $F_t$, surprise is:
\begin{equation}
\label{eq:surprise}
\mathrm{surprise}(h_t,F_t) = |F_t - \hat{F}_t|.
\end{equation}
Replay samples are drawn with probability proportional to
surprise$^\alpha$:
\begin{equation}
\label{eq:surprise-replay}
P(\mathrm{replay}_i)
\propto
\mathrm{surprise}_i^{\alpha},
\qquad \alpha = 0.6.
\end{equation}
The purpose of surprise-driven replay is to focus the reasoner on
campaign outcomes that contradict its current expectations.

\subsection{Scientific Reasoner}
\label{app:mscl_reasoner}

The Scientific Reasoner is a small LM $\phi$ that proposes the next
strategy:
\begin{equation}
h_{t+1} = \phi(M_L, M_M, F_t, \mathrm{ctx}_{t+1}),
\end{equation}
where $M_L$ and $M_M$ are the long- and mid-term memories, $F_t$ is the
most recent observed performance, and $\mathrm{ctx}_{t+1}$ is the
context for the next campaign. The reasoner emits both a strategy
$h_{t+1}$ and a natural-language rationale.

\paragraph{Scientific DPO loss.}
Unlike standard DPO, where preference pairs usually come from human or
AI feedback over responses, MSCL derives preference pairs from
experimental outcomes. Given two candidate strategies $h_w$ and $h_l$,
we prefer $h_w$ over $h_l$ if:
\[
F_t(h_w) > F_t(h_l) + \varepsilon.
\]
The reasoner is trained with a DPO-style objective:
\begin{equation}
\label{eq:scidpo}
\begin{aligned}
\mathcal{L}_{\mathrm{SciDPO}}
=
-\mathbb{E}\Big[
\log\sigma\big(
&\beta_s
\log
\tfrac{\phi(h_w|x)}{\phi_{\mathrm{ref}}(h_w|x)}
\\[-1pt]
&-
\beta_s
\log
\tfrac{\phi(h_l|x)}{\phi_{\mathrm{ref}}(h_l|x)}
\big)
\Big],
\end{aligned}
\end{equation}
where $x$ denotes the memory and campaign context, $\phi_{\mathrm{ref}}$
is the reference reasoner, and $\beta_s$ controls the preference
optimization strength.

\paragraph{EWC regularization.}
To reduce overwriting of validated principles, MSCL adds an EWC-style
regularization term:
\begin{equation}
\label{eq:ewc}
\mathcal{L}_{\mathrm{total}}
=
\mathcal{L}_{\mathrm{SciDPO}}
+
\tfrac{\lambda}{2}
\sum_i
F_i
(\phi_i - \phi_i^\star)^2.
\end{equation}
Here, $\phi_i^\star$ denotes the parameter value after learning prior
validated principles, and $F_i$ is the Fisher-style importance weight for
parameter $i$. This term penalizes changes to parameters associated with
previously validated methodological knowledge.

\paragraph{Interpretation.}
MSCL is designed to test whether a memory-augmented reasoner can
accumulate principle-level training knowledge across campaigns. The main
paper shows that this more expressive mechanism does not automatically
solve scientific amnesia: it improves only in the completed
heterogeneous-chain pilot, while a simpler retrieval-only baseline has
the best mean $\Delta$ in the small homogeneous multi-seed sweep. This is
why the paper treats MSCL as a diagnostic candidate rather than as a
claimed solution.

\section{Synthetic Mock Pilot}
\label{app:mock}

\begin{table*}[t]
\centering
\caption{Mock pilot, 5 seeds: mean pass@1 with two-sided Mann--Whitney U vs.\ MSCL Reasoner.}
\label{tab:mock_pilot_multiseed}
\begin{tabular}{lrrr}
\toprule
Condition & Mean pass@1 & Total regret & Mean surprise \\
\midrule
Random Memory & 0.772$\pm$0.009* & 6.84$\pm$0.26 & 0.117$\pm$0.008 \\
Rule-based & 0.767$\pm$0.077* & 6.98$\pm$2.32 & 0.082$\pm$0.010 \\
Retrieval-only & 0.771$\pm$0.076* & 6.88$\pm$2.29 & 0.085$\pm$0.012 \\
Warm-start BO & 0.864$\pm$0.011 & 4.09$\pm$0.34 & 0.110$\pm$0.008 \\
MSCL Reasoner & 0.862$\pm$0.007 & 4.13$\pm$0.21 & 0.109$\pm$0.009 \\
\bottomrule
\end{tabular}
\end{table*}

\paragraph{Comparative intervention readout.}
On the synthetic landscape, MSCL attains mean pass@1
$=\mathbf{0.862 \pm 0.007}$, exceeding three of the other four
candidates: B7 Random Memory ($0.772 \pm 0.009$, $p{=}0.012$),
B6 Rule-based ($0.767 \pm 0.077$, $p{=}0.012$), and
B4 Retrieval-only ($0.771 \pm 0.076$, $p{=}0.022$); it is
indistinguishable from B2 Warm-start BO
($0.864 \pm 0.011$, $p{=}0.53$). This is algorithmic validation on a
synthetic landscape that rewards UCB-style exploration, not evidence
that any candidate is superior on real-LM continual DPO.

\begin{figure}[t]\centering
  \includegraphics[width=0.6\linewidth]{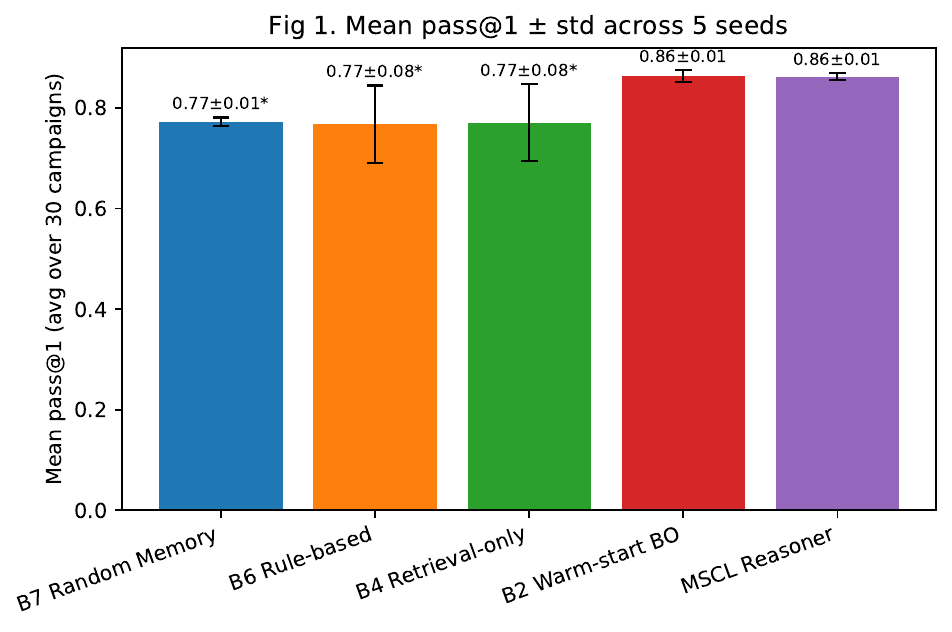}
  \caption{Mock pilot: mean pass@1 $\pm$ std across 5 seeds.
  Significance stars are two-sided Mann--Whitney U vs.\ MSCL, treated
  here as the fixed comparison candidate.}
  \label{fig:mock-pass}
\end{figure}

\begin{figure}[t]\centering
  \includegraphics[width=0.6\linewidth]{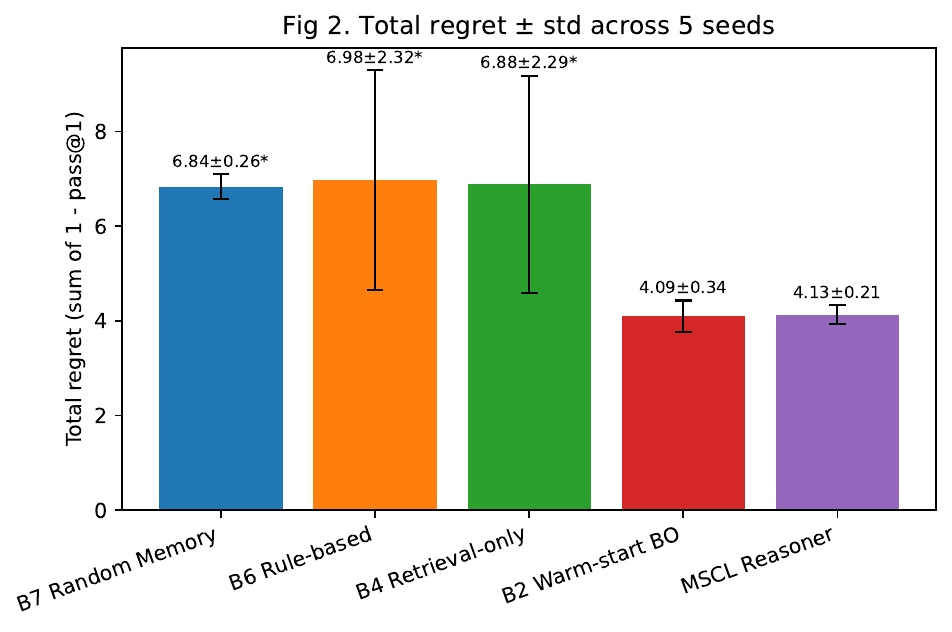}
  \caption{Mock pilot: cumulative regret across the 30-campaign
  sequence, mean $\pm$ std over 5 seeds.}
  \label{fig:mock-regret}
\end{figure}

\section{Single-Step Real-LM Pilot}
\label{app:single_step}

Each of the five candidate proposers submits one MAST job on the
cold-start \texttt{humaneval\_control\_easy} subdomain. All five jobs
reach \texttt{COMPLETE exit\_code=0}. We harvest the per-batch rolling
pass@1 at the first and last training steps and the peak across 50
steps.

\begin{table*}[t]
\centering
\caption{Real-LM single-step pilot on Qwen2.5-7B-Instruct (M2 lite v2). Each condition: 1 campaign on \texttt{humaneval\_control\_easy}, 50 train steps; \texttt{pass@1/mean} TB tag from CodeGrader. Peak = max across 50 steps.}
\label{tab:real_lm_m2_lite_v2}
\begin{tabular}{lrrrr}
\toprule
Condition & Strategy & Initial & Final & Peak \\
\midrule
Random Memory   & $\beta{=}0.1,\,\eta{=}10^{-4},\,T{=}50$  & 0.062 & 0.312 & 0.438 \\
Rule-based      & $\beta{=}0.1,\,\eta{=}10^{-4},\,T{=}200$ & 0.062 & \textbf{0.438} & 0.438 \\
Retrieval-only  & $\beta{=}0.1,\,\eta{=}10^{-4},\,T{=}200$ & 0.188 & 0.188 & \textbf{0.562} \\
Warm-start BO   & $\beta{=}0.1,\,\eta{=}10^{-4},\,T{=}200$ & 0.125 & 0.250 & 0.375 \\
MSCL Reasoner      & $\beta{=}0.1,\,\eta{=}5{\cdot}10^{-5},\,T{=}200,\,bs{=}2$ & 0.062 & 0.000 & 0.500 \\
\bottomrule
\end{tabular}
\end{table*}

All five candidates improve at their peak step. End-of-run trajectories
diverge: B6 Rule-based converges and stays at 0.438, while B4
Retrieval-only and MSCL reach higher peaks but degrade by the final
step. We use this pilot to validate real-LM execution and evaluator
behavior; chain-level scientific-amnesia claims are based on the
multi-step experiments in the main text.
\section{Pipeline and Chain Execution Details}
\label{app:pipeline_results}

Multi-step real-LM chains are launched by
\texttt{infra/mast/chain\_launcher.py}. For each campaign step, the
launcher submits one MAST job under the
\texttt{v0.mscl\_chain\_step} configuration, polls the job to a terminal
state, discovers the FSDP-sharded trainer save via
\texttt{find\_final\_checkpoint}, and passes that path into the next job
as \texttt{--cluster\_config.model\_name}. This implements the
checkpoint-passthrough behavior used in all real-LM chain experiments.

The initial real-LM validation has 20/20 completed MAST jobs: five
single-step jobs and fifteen homogeneous-chain jobs. In Block~D, two of
five heterogeneous chains did not complete because the Python chain
driver exited before the relevant MAST job reached \texttt{COMPLETE}. In
Block~B, two of fifteen condition-seed chains are missing for the same
reason. These incomplete chains are treated as robustness gaps rather
than evidence for or against any intervention.
\end{document}